\documentclass[11pt,a4paper]{article}

\usepackage[T1]{fontenc}
\usepackage[utf8]{inputenc}
\usepackage{mathpazo}
\usepackage{microtype}

\usepackage[a4paper,margin=2.5cm]{geometry}
\usepackage{amsmath,amssymb}

\usepackage{graphicx}
\usepackage{booktabs}
\usepackage{array}
\usepackage{tabularx}
\usepackage{multirow}
\usepackage{ragged2e}
\usepackage{float}
\usepackage{placeins}
\usepackage{caption}
\usepackage{enumitem}
\setlist[itemize,enumerate]{
  leftmargin=*,
  itemsep=0.2em,
  topsep=0.4em
}

\usepackage{tikz}
\usetikzlibrary{positioning,calc,arrows.meta}

\usepackage{authblk}

\usepackage[numbers,sort&compress]{natbib}

\usepackage{xcolor}
\definecolor{linkblue}{RGB}{0,70,120}

\usepackage{hyperref}
\hypersetup{
  colorlinks=true,
  linkcolor=linkblue,
  citecolor=linkblue,
  urlcolor=linkblue,
  pdftitle={Learning and Structurally Validating Simulation Scenario Continuations in Dynamic Graph Systems},
  pdfauthor={Michael Romei de Socio, Gian Luca Pozzato, Alessio Merlo}
}

\usepackage{orcidlink}
\usepackage[capitalise,nameinlink,noabbrev]{cleveref}

\title{
  Learning and Structurally Validating Simulation Scenario Continuations in Dynamic Graph Systems
}

\author[1,2]{
  Michael Romei de Socio\,
  \orcidlink{0009-0008-3949-0437}
}

\author[2]{
  Gian Luca Pozzato\,
  \orcidlink{0000-0002-3952-4624}
}

\author[1]{
  Alessio Merlo\,
  \orcidlink{0000-0002-2272-2376}
  \thanks{
    Corresponding author:
    \href{mailto:alessio.merlo@unicasd.it}
    {alessio.merlo@unicasd.it}; Tel.: +39 06 469123191
  }
}

\affil[1]{
  CASD -- School of Advanced Defense Studies,
  00165 Rome, Italy
}

\affil[2]{
  Department of Computer Science, University of Turin,
  10149 Turin, Italy
}

\date{\small Preprint -- August 2026}

\begin{document}

\maketitle

\begin{abstract}
Data-driven generative models can extend partially observed simulation trajectories into ensembles of alternative future scenarios. However, consistency with a learned trajectory distribution does not guarantee that a generated continuation satisfies the structural conditions of the simulated system. This paper presents a method for learned scenario continuation and post-generation structural validation in dynamic graph simulations. A conditional diffusion model generates future graph-state trajectories from partial simulation histories, while an external symbolic layer evaluates each continuation through a Boolean admissibility indicator and a continuous violation score. The resulting information supports two primary scenario-management strategies, hard filtering and soft weighting; optional projection is additionally considered as an illustrative deterministic repair baseline.

The method is evaluated on two controlled dynamic-graph simulation regimes that share the same continuation architecture and training protocol but differ in dimensionality, dependency complexity, and selected regime-specific parameters. Evaluation considers invalid probability mass, scenario retention, effective sample size, diversity, robustness, and probability calibration. In the compact positive-control regime, unconstrained invalid mass is 0.002996, indicating that the learned continuation distribution is almost entirely contained within the admissible scenario space. In the medium-complexity regime, invalid mass increases to 0.155929. Hard filtering removes all invalid scenarios from the retained ensemble while preserving 84.4\% of the generated continuations. Soft weighting retains an effective sample size ratio of 0.998764 but reduces invalid mass only to 0.148807. For comparison, optional projection reduces invalid mass to 0.000023 by directly modifying the generated trajectories. Dependency constraints account for nearly all observed inadmissibility.

The results show a substantial divergence between the learned continuation distribution and the admissible scenario space in the structurally richer regime. Learned-distribution support, structural admissibility, and probability calibration should therefore be assessed separately when learned components are used for simulation-scenario generation and management.
\end{abstract}


\bigskip

\section{Introduction}
\label{sec:introduction}

Simulation supports the analysis of complex systems by enabling exploration of alternative evolutions, stress conditions, recovery processes, and interactions that cannot be examined exhaustively in the real system. In many application domains, the simulated system is naturally represented as a dynamic graph: nodes encode component states, while directed edges describe functional, operational, or information dependencies \cite{kazemi2020representation,jin2024spatiotemporal}. Multilayer and interdependent structures further allow local changes to propagate through coupled components and produce system-level effects that cannot be understood from individual state variables alone \cite{kivela2014multilayer,buldyrev2010catastrophic}.

Data-driven and surrogate simulation methods use learned approximations to reproduce or extend the behavior of computational models, often to reduce the cost of repeated model evaluations or support many-query analyses \cite{forrester2008surrogate}. Their assessment commonly considers predictive fidelity, computational efficiency, and uncertainty representation. When a learned component generates complete scenario continuations, however, local predictive quality does not by itself establish consistency with the structural semantics of the reference simulation. This distinction complements established simulation verification and validation principles, which require a model and its results to be assessed with respect to their intended use and domain of applicability \cite{sargent2013verification}. It motivates treating structural validation of generated scenarios as a separate element of a learned simulation workflow.

A central simulation task is the construction of alternative future scenarios from an observed system history. Classical simulators typically generate these scenarios from explicit transition rules, calibrated stochastic processes, or domain models. Data-driven generative models provide a complementary mechanism: they can learn a conditional distribution over future trajectories and produce multiple scenario continuations from the same partially observed state. Diffusion models, score-based models, normalizing flows, neural differential equations, and related approaches have substantially expanded the range of high-dimensional temporal distributions that can be learned and sampled \cite{ho2020denoising,song2021score,song2021denoising,papamakarios2021normalizing,kobyzev2021normalizing,chen2018neural,lipman2023flow}.

This capability is attractive for simulation because it supports ensemble-based exploration rather than a single predicted evolution. Given a common observation history, a learned generator can produce multiple continuations that represent alternative future realizations. Such ensembles can expose uncertainty, rare evolutions, and heterogeneous system responses. However, statistical consistency with the learned trajectory distribution is not sufficient to establish the admissibility of a generated simulation scenario. A continuation may resemble the training data while violating structural properties of the modeled system, including dependency relations, prerequisite conditions, mutual exclusions, temporal orderings, or graph-state invariants.

We refer to this distinction as the gap between \emph{statistical plausibility} and \emph{structural admissibility}. Statistical plausibility concerns whether a generated trajectory is supported by the learned conditional distribution. Structural admissibility concerns whether that trajectory satisfies the explicit relations and constraints defining allowable system behavior. The two properties are complementary but not equivalent. A statistically plausible continuation may be structurally inadmissible, while a structurally admissible continuation may still be poorly represented or assigned an unreliable probability by the learned model.

This distinction has direct implications for data-driven simulation. If learned scenarios are evaluated only through predictive or distributional criteria, structurally impossible continuations may remain in the scenario ensemble and influence subsequent analyses. Conversely, enforcing admissibility by rejecting or modifying scenarios may reduce sample efficiency, alter the generated distribution, or suppress scenario diversity. Structural validation is therefore not merely a post-processing detail: it creates an explicit trade-off between the fidelity, admissibility, and usability of generated simulation scenarios.

The problem becomes particularly relevant as structural complexity increases. In a compact graph, a generative model may learn most local and relational regularities directly from the available trajectories. In a graph with multiple parents, hierarchical outputs, cross-layer couplings, and temporal prerequisites, locally plausible values must also remain jointly consistent across interconnected components. Increasing structural complexity may therefore widen the gap between what the generator considers statistically plausible and what the simulation model defines as admissible.

This paper investigates this gap across two regimes of structural complexity through a method for learned scenario continuation and post-generation structural validation in dynamic graph systems. Starting from a partial observation window $O_t$ and a latent variable $z$, the method generates a future graph-state trajectory through the conditional operator

\begin{equation}
\tau = T_{\theta}(z,O_t).
\label{eq:intro_conditional_flow_map}
\end{equation}

Throughout the paper, the term \emph{flow map} is used in the dynamical-systems sense to denote the learned conditional operator that maps an observed system history and latent sampling uncertainty into a future trajectory. In the diffusion-based implementation considered here, $z$ represents the complete source of sampling randomness, including the initial diffusion noise and, when applicable, stochasticity introduced during reverse sampling. Accordingly, $T_{\theta}$ is interpreted as a learned conditional scenario-continuation operator, rather than as a deterministic simulator or as a replacement for the underlying simulation model.

The proposed approach separates scenario generation from scenario validation. A conditional diffusion model first generates an ensemble of future graph-state continuations. An external symbolic layer then assesses each generated scenario through a Boolean admissibility indicator and a continuous violation score. The primary post-generation strategies are hard filtering, which restricts the ensemble to admissible scenarios, and soft weighting, which preserves all scenarios but redistributes their probability mass according to violation severity. Optional projection is additionally considered as an illustrative deterministic repair baseline for the distinct case in which generated trajectories are directly modified. This separation makes it possible to analyze the effect of structural validation without embedding the constraints in the generator or changing its training objective.

Constraints could alternatively be incorporated during training or used to guide the generative process itself \cite{chung2023diffusion,christopher2024constrained}. The present study deliberately focuses on post-generation validation because it preserves a clear boundary between the learned scenario model and the explicit structural knowledge of the simulated system. This modularity is useful when constraints are maintained separately from the data-driven generator, when different validation policies must be compared, or when an existing generative component must be integrated into a simulation workflow without retraining it.

The study addresses two research questions:

\begin{enumerate}
\item How does the divergence between statistical plausibility and structural admissibility differ across the two graph-structural complexity regimes?

\item How do hard filtering and soft weighting trade structural admissibility against scenario retention, effective sample size, diversity, robustness, and probability calibration?
\end{enumerate}

To answer these questions, we construct two controlled dynamic-graph simulation regimes that share the same continuation architecture, training protocol, and evaluation pipeline, but differ in state dimensionality, dependency complexity, and selected regime-specific parameters. The compact regime acts as a positive control in which most structural regularities can be learned from the simulated trajectories. The medium-complexity regime introduces multi-parent dependencies, hierarchical service outputs, cross-layer coupling, and temporal constraints. This design enables a controlled comparison of learned scenario admissibility across two levels of structural complexity while holding the continuation architecture and training protocol fixed.

The contributions of this paper are:

\begin{enumerate}
\item a flow-map formulation of learned scenario continuation for partially observed dynamic graph simulations;

\item a conditional diffusion method for generating ensembles of future simulation scenarios from common observation histories;

\item a modular symbolic validation layer that distinguishes statistical plausibility from structural admissibility and supports filtering and reweighting, together with optional projection as a deterministic repair baseline;

\item a controlled two-regime simulation benchmark for comparing the admissibility of learned scenario continuations across different levels of structural complexity;

\item a multidimensional evaluation of scenario admissibility and ensemble usability, covering invalid probability mass, retention, effective sample size, diversity, robustness, and calibration.
\end{enumerate}

The remainder of the paper is organized as follows. Section~\ref{sec:materials_methods} presents the learned scenario-continuation framework, the structural validation strategies, and the controlled simulation benchmark. Section~\ref{sec:results} reports the experimental results. Section~\ref{sec:discussion} discusses their implications for data-driven simulation and scenario management. Section~\ref{sec:conclusions} summarizes the main findings and outlines future research directions.

\section{Methodology}
\label{sec:materials_methods}

\subsection{Learned Scenario-Continuation Framework}
\label{subsec:methods_framework}

The proposed framework augments a simulation workflow with three components: a learned scenario-continuation model, a structural validation layer, and a scenario-management layer.

A controlled stochastic simulator first produces the reference trajectories from which the conditional dynamics are learned. Given a partial observation history $O_t$ and latent sampling uncertainty $z$, the learned model then generates alternative continuations of the observed scenario according to the conditional flow map introduced in Equation~\eqref{eq:intro_conditional_flow_map}. The generator approximates the conditional distribution
\begin{equation}
P_{\theta}(\tau \mid O_t),
\label{eq:conditional_distribution}
\end{equation}
where $\tau$ denotes a future dynamic-graph trajectory.

Each generated continuation is evaluated by an external symbolic layer through a Boolean admissibility indicator $I_C(\tau)$ and a continuous violation score $\phi_C(\tau)$. These quantities support two primary scenario-management strategies: hard filtering and soft weighting. Optional projection is evaluated separately as an illustrative deterministic repair baseline in which generated trajectories are directly modified. The resulting ensemble can be retained as a validated set of alternative simulation scenarios or used to estimate scenario-level outcome probabilities.

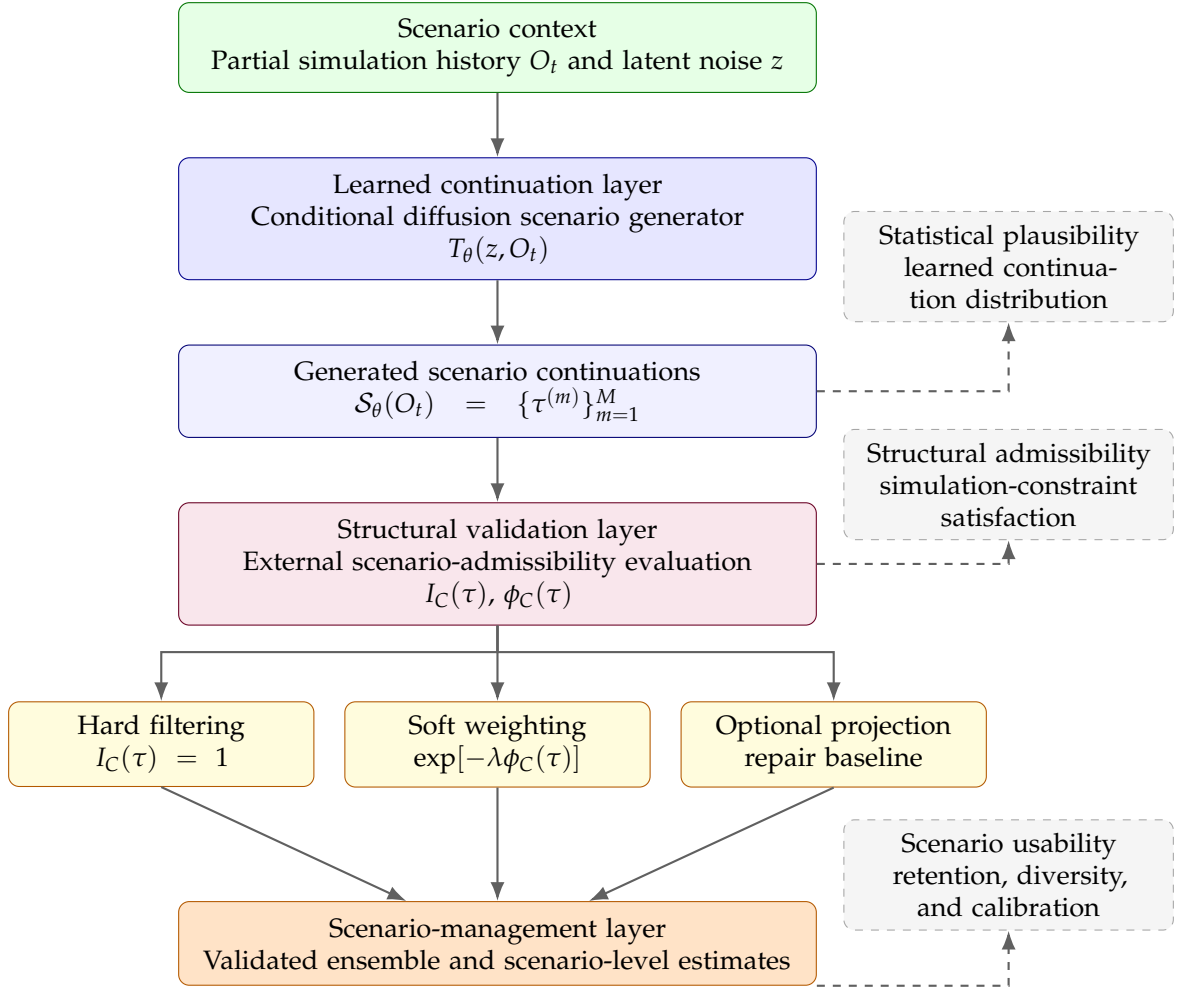
\begin{figure}[!t]
\centering
\begin{tikzpicture}[
font=\small,
node distance=0.85cm,
mainbox/.style={draw,rounded corners,align=center,text width=0.50\textwidth,minimum height=0.95cm,inner sep=6pt},
inputbox/.style={mainbox,fill=green!10,draw=green!45!black},
neuralbox/.style={mainbox,fill=blue!10,draw=blue!55!black},
trajbox/.style={mainbox,fill=blue!6,draw=blue!45!black},
constraintbox/.style={mainbox,fill=purple!10,draw=purple!55!black},
decisionbox/.style={mainbox,fill=orange!22,draw=orange!70!black},
strategybox/.style={draw,rounded corners,align=center,text width=0.23\textwidth,minimum height=0.85cm,inner sep=5pt,fill=yellow!16,draw=orange!70!black},
relbox/.style={draw,dashed,rounded corners,align=center,text width=0.25\textwidth,minimum height=0.85cm,inner sep=5pt,fill=gray!8,draw=gray!70},
arrow/.style={-{Latex[length=2.5mm]},thick,draw=gray!75!black}
]
\node[inputbox] (inputs)
{Scenario context\\Partial simulation history $O_t$ and latent noise $z$};
\node[neuralbox, below=of inputs] (generator)
{Learned continuation layer\\Conditional diffusion scenario generator\\$T_{\theta}(z,O_t)$};
\node[trajbox, below=of generator] (traj)
{Generated scenario continuations\\$\mathcal{S}_{\theta}(O_t)=\{\tau^{(m)}\}_{m=1}^{M}$};
\node[constraintbox, below=of traj] (constraints)
{Structural validation layer\\External scenario-admissibility evaluation\\$I_C(\tau),\,\phi_C(\tau)$};
\node[strategybox, below left=1cm and -1.8cm of constraints] (hard)
{Hard filtering\\$I_C(\tau)=1$};
\node[strategybox, below=1.0cm of constraints] (soft)
{Soft weighting\\$\exp[-\lambda\phi_C(\tau)]$};
\node[strategybox, below right=1.0cm and -1.8cm of constraints] (proj)
{Optional projection\\repair baseline};
\node[decisionbox, below=1.5cm of soft] (decision)
{Scenario-management layer\\Validated ensemble and scenario-level estimates};
\node[relbox, below right=-3.0cm and 0.35cm of traj] (plausibility)
{Statistical plausibility\\learned continuation distribution};
\node[relbox, below right=-2.6cm and 0.35cm of constraints] (admissibility)
{Structural admissibility\\simulation-constraint satisfaction};
\node[relbox, below right=-2.2cm and 0.35cm of decision] (usability)
{Scenario usability\\retention, diversity, and calibration};
\draw[arrow] (inputs) -- (generator);
\draw[arrow] (generator) -- (traj);
\draw[arrow] (traj) -- (constraints);
\draw[arrow] (constraints.south) -- ++(0,-0.35) -| (hard.north);
\draw[arrow] (constraints.south) -- (soft.north);
\draw[arrow] (constraints.south) -- ++(0,-0.35) -| (proj.north);
\draw[arrow] (hard.south) -- (decision);
\draw[arrow] (soft.south) -- (decision.north);
\draw[arrow] (proj.south) -- (decision);
\draw[arrow, dashed] (traj.east) -| (plausibility.south);
\draw[arrow, dashed] (constraints.east) -| (admissibility.south);
\draw[arrow, dashed] (decision.south east) -| (usability.south);
\end{tikzpicture}
\caption{Constraint-aware learned scenario-continuation framework. A conditional diffusion model extends a partial simulation history into an ensemble of future dynamic-graph scenarios. An external structural layer evaluates scenario admissibility and supports hard filtering and soft weighting, while optional projection is included as a deterministic repair baseline before subsequent scenario analysis.}
\label{fig:framework}
\end{figure}

The architecture is summarized in Figure~\ref{fig:framework}. Its modular structure separates four functions that are often conflated in data-driven simulation: reference-trajectory generation, learned scenario continuation, structural validation, and downstream scenario analysis. This separation makes it possible to evaluate statistical plausibility, structural admissibility, ensemble usability, and probability calibration independently.

\subsection{Dynamic-Graph Simulation State}
\label{subsec:methods_dynamic_graphs}

Each simulated system state is represented as a dynamic directed graph
\begin{equation}
G_t=(V,E,X_t,A_t),
\label{eq:dynamic_graph}
\end{equation}
where $V=\{1,\ldots,n\}$ is the node set, $E \subseteq V\times V$ is the set of allowed directed dependencies, $X_t\in[0,1]^n$ contains node states, and $A_t\in[0,1]^{|E|}$ contains allowed-edge states. Node values represent operational, functional, or integrity conditions, while edge values describe the evolving state of admissible dependencies. Forbidden edges are excluded from the generated state representation.

The graph-state vector is
\begin{equation}
s_t=[X_t,A_t]\in [0,1]^d,
\qquad d=n+|E|,
\label{eq:graph_state_vector}
\end{equation}
and the available observation history is
\begin{equation}
O_t=\{\tilde{s}_{t-W+1},\ldots,\tilde{s}_{t}\},
\label{eq:partial_observation}
\end{equation}
where $\tilde{s}_t$ may be noisy or partially observed. The scenario-continuation target is
\begin{equation}
\tau_{t+1:t+H}=\{s_{t+1},\ldots,s_{t+H}\}\in[0,1]^{H\times d},
\label{eq:future_trajectory}
\end{equation}
which is denoted simply by $\tau$ throughout the paper.

In the medium-complexity regime, $K_1$ represents an intermediate local service outcome, whereas $K_2$ represents the final global service outcome. These variables provide scenario-level observables through which alternative continuations can be compared and probabilistically summarized. The admissible high-service event associated with $K_2$ is retained as a calibration target, but it is not the sole purpose of the generated ensemble. Allowed-edge variables are generated jointly with node states and evaluated through boundedness and continuity-related invariants, while dependency constraints are defined at the node level.

\subsection{Conditional Diffusion for Scenario Continuation}
\label{subsec:methods_conditional_diffusion}

The learned continuation model approximates the distribution of future simulation trajectories conditional on the available scenario history, $P_{\theta}(\tau\mid O_t)$. We instantiate this model using a lightweight conditional diffusion architecture. Training follows the standard noise-prediction objective
\begin{equation}
\mathcal{L}_{\mathrm{diff}}(\theta)
=
\mathbb{E}_{\tau,O_t,\epsilon,k}
\left[
\left\|\epsilon-\epsilon_{\theta}(\tau_k,k,O_t)\right\|_2^2
\right],
\label{eq:diffusion_loss}
\end{equation}
where $\epsilon\sim\mathcal N(0,I)$ and
\begin{equation}
\tau_k
=
\sqrt{\bar{\alpha}_k}\,\tau
+
\sqrt{1-\bar{\alpha}_k}\,\epsilon,
\label{eq:forward_diffusion_trajectory}
\end{equation}
with $\bar{\alpha}_k$ denoting the cumulative noise-schedule coefficient.

At inference time, the model generates an ensemble
\begin{equation}
\mathcal S_{\theta}(O_t)=\{\tau^{(1)},\ldots,\tau^{(M)}\},
\qquad
\tau^{(m)}\sim P_{\theta}(\tau\mid O_t).
\label{eq:sample_ensemble}
\end{equation}
For a common observed history, repeated sampling produces an ensemble of alternative scenario continuations. The model therefore acts as a stochastic continuation component within the simulation workflow, rather than as a one-step predictor or a replacement for the reference simulator.

The generator receives simulated observations, observation masks, and diffusion-step embeddings, but never receives symbolic constraints, validity labels, or violation scores. This separation prevents constraint leakage and allows post-generation structural validation to be evaluated independently. Generated continuous outputs are mapped component-wise to $[0,1]$ before constraint evaluation.

\subsection{Structural Scenario-Validation Layer}
\label{subsec:methods_symbolic_constraints}

Let $C$ denote the set of structural conditions defining the admissible behavior of the simulated system. Each generated scenario continuation receives a Boolean admissibility indicator
\begin{equation}
I_C(\tau)=
\begin{cases}
1, & \text{if } \tau \text{ satisfies all constraints in } C,\\
0, & \text{otherwise,}
\end{cases}
\label{eq:validity_indicator}
\end{equation}
and a continuous violation score
\begin{equation}
\phi_C(\tau)\ge0,
\label{eq:violation_score}
\end{equation}
where $\phi_C(\tau)=0$ for fully admissible scenarios and larger values indicate more severe violations.

Hard filtering restricts the generated ensemble to structurally admissible scenario continuations:
\begin{equation}
\mathcal S_{\mathrm{hard}}(O_t,C)
=
\{\tau^{(m)}\in\mathcal S_{\theta}(O_t):I_C(\tau^{(m)})=1\}.
\label{eq:hard_filtered_set}
\end{equation}
When the valid mass is positive, this induces
\begin{equation}
P_{\mathrm{valid}}(\tau\mid O_t,C)=
\frac{P_{\theta}(\tau\mid O_t)I_C(\tau)}{Z_{\mathrm{valid}}(O_t,C)},
\qquad
Z_{\mathrm{valid}}(O_t,C)=
\int P_{\theta}(\tau\mid O_t)I_C(\tau)\,d\tau.
\label{eq:hard_filter_distribution}
\end{equation}
Hard filtering provides an explicit admissibility guarantee for the retained ensemble, at the cost of discarding generated scenarios.

Soft weighting retains every generated continuation but reduces the contribution of scenarios with larger violation scores:
\begin{equation}
w^{(m)}=
\frac{\exp[-\lambda\phi_C(\tau^{(m)})]}
{\sum_{j=1}^{M}\exp[-\lambda\phi_C(\tau^{(j)})]}.
\label{eq:soft_weights}
\end{equation}
The corresponding weighted distribution is
\begin{equation}
P_{\mathrm{soft}}(\tau\mid O_t,C)=
\frac{P_{\theta}(\tau\mid O_t)\exp[-\lambda\phi_C(\tau)]}{Z_{\mathrm{soft}}(O_t,C)},
\qquad
Z_{\mathrm{soft}}(O_t,C)=
\int P_{\theta}(\tau\mid O_t)\exp[-\lambda\phi_C(\tau)]\,d\tau.
\label{eq:soft_weight_distribution}
\end{equation}
This violation-dependent exponential penalty is related to knowledge-based regularization in generative modeling \cite{takeishi2020knowledge}, but is applied here only after sampling. Soft weighting therefore changes the probability distribution over the scenario ensemble without guaranteeing Boolean admissibility.

Optional projection is included as an illustrative deterministic repair baseline. It applies a fixed rule-based map $\Pi_C:\tau\mapsto\tilde{\tau}$ to each generated continuation. The term ``projection'' is used here as shorthand for this repair transformation and does not denote an optimization-based metric projection onto the admissible set. Node and allowed-edge values are first clipped to $[0,1]$. A single sequential sweep is then applied to the node states using the same dependency slack $\eta=0.05$ adopted by the structural validator. In the compact regime, the dependency caps for $C$, $D$, and $K$ are applied in this order, followed by the maintenance--service exclusion for $K$. In the medium-complexity regime, the corresponding caps are applied sequentially to $C_1$, $C_2$, $D_1$, $D_2$, $Q$, $K_1$, and $K_2$, followed by the maintenance--service exclusions for $K_1$ and $K_2$. Finally, consecutive allowed-edge changes are limited to the maximum admissible jump $\Delta_{\max}=0.25$.

The repaired trajectory is returned after this single sweep: no iterative refinement, feasibility search, convergence criterion, or stopping rule is used. Temporal-ordering constraints are not explicitly repaired, and later updates within the sequential sweep may modify quantities used by earlier dependency caps. The procedure therefore does not guarantee exact admissibility. Because it directly modifies generated scenario content, optional projection is treated as an illustrative repair baseline rather than as a third validation strategy equivalent to hard filtering or soft weighting.

Five constraint families are considered: threshold, dependency, mutual-exclusion, temporal-ordering, and structural-invariant constraints. They represent complementary properties of scenario admissibility and are not mutually exclusive; a generated continuation may violate multiple families simultaneously. The constraint families are formalized in Appendix~\ref{app:constraint_families_violation_scores}, while complete regime-specific specifications are provided in the Supplementary Material.

\subsection{Controlled Simulation Benchmark}
\label{subsec:methods_two_regime_benchmark}

The experimental study uses two controlled dynamic-graph simulation regimes implemented within the same stochastic simulation framework and evaluated through the same learned-continuation and validation pipeline. They use the same continuation architecture and training protocol, while retaining regime-specific simulator and sampling parameters. The regimes differ principally in graph dimensionality and dependency complexity.

Experiment~1 is a compact positive-control regime in which the learned model is expected to capture most local and relational regularities. Experiment~2 is a medium-complexity regime introducing richer multi-parent dependencies, hierarchical service variables, cross-layer coupling, and temporal constraints. The paired design therefore provides a controlled comparison of learned scenario admissibility across two levels of structural complexity while holding the continuation architecture, training protocol, prediction horizon, ensemble size, and evaluation procedure fixed.

\begin{figure}[tp]
\centering
\includegraphics[width=0.95\textwidth]{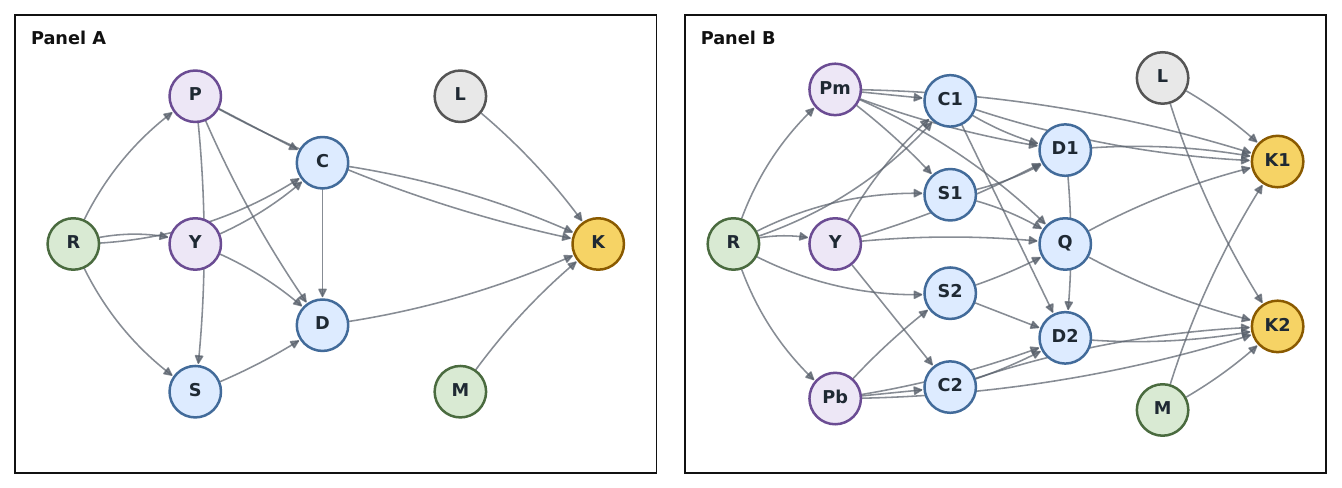}
\caption{Dynamic-graph simulation regimes. Panel A shows the compact graph with 9 nodes and 16 allowed directed edges; Panel B shows the medium-complexity graph with 15 nodes and 37 allowed directed edges.}
\label{fig:graph_topologies}
\end{figure}

The benchmark specifications are summarized in Table~\ref{tab:benchmark_specification}, and the corresponding graph topologies are shown in Figure~\ref{fig:graph_topologies}.

\begin{table}[htbp]
\caption{Controlled dynamic-graph simulation regimes used to evaluate learned scenario continuation and structural validation. The regimes share the same overall simulation framework, continuation architecture, training protocol, and evaluation procedure, but differ in graph size, state dimensionality, dependency complexity, and selected regime-specific parameters.}
\label{tab:benchmark_specification}
\centering
\small
\setlength{\tabcolsep}{5pt}
\renewcommand{\arraystretch}{1.08}
\begin{tabularx}{\linewidth}{@{}>{\raggedright\arraybackslash}p{0.44\linewidth}
>{\centering\arraybackslash}p{0.24\linewidth}
>{\centering\arraybackslash}p{0.24\linewidth}@{}}
\toprule
\textbf{Quantity} &
\shortstack{\textbf{Experiment 1}\\\textbf{compact}} &
\shortstack{\textbf{Experiment 2}\\\textbf{complex}} \\
\midrule
\multicolumn{3}{@{}l}{\textit{Graph and target specification}} \\
\midrule
Nodes & 9 & 15 \\
Allowed directed edges & 16 & 37 \\
Graph-state dimension ($d=n+|E|$) & 25 & 52 \\
Prediction horizon ($H$) & 16 & 16 \\
Flattened target dimension ($H \times d$) & 400 & 832 \\
\addlinespace[0.3em]
\multicolumn{3}{@{}l}{\textit{Dataset construction and sampling}} \\
\midrule
Base trajectories & 3000 & 3000 \\
Trajectory length & 80 & 80 \\
Observation window ($W$) & 8 & 8 \\
Train/validation/test split & 70/15/15 & 70/15/15 \\
Test contexts per seed & 450 & 450 \\
Samples per context & 512 & 512 \\
\addlinespace[0.3em]
\multicolumn{3}{@{}l}{\textit{Replication}} \\
\midrule
Random seeds & $2026$, $2027$, $2028$ & $2026$, $2027$, $2028$ \\
\bottomrule
\end{tabularx}
\end{table}

\subsubsection{Experiment 1: Compact Graph Regime}
\label{subsubsec:methods_experiment_1}

Experiment~1 contains nine nodes,
\begin{equation}
V_{\mathrm{compact}}=\{P,Y,C,S,D,L,R,M,K\},
\label{eq:compact_nodes}
\end{equation}
representing power availability, cyber integrity, communication reliability, sensing coverage, data-processing integrity, logistics readiness, repair capacity, maintenance or safe mode, and service feasibility. The graph includes 16 allowed directed dependency edges, yielding $d=25$ and a flattened target dimension of $H\times d=400$. This regime acts as a positive control for evaluating whether learned scenario continuations remain admissible under limited structural complexity.

\subsubsection{Experiment 2: Medium-Complexity Dependency Graph Regime}
\label{subsubsec:methods_experiment_2}

Experiment~2 contains 15 nodes:
\begin{equation}
\begin{split}
V_{\mathrm{complex}}=
\{&P_m,P_b,Y,C_1,C_2,S_1,S_2,D_1,D_2,Q,\\
&L,R,M,K_1,K_2\}.
\end{split}
\label{eq:complex_nodes}
\end{equation}
The graph contains 37 allowed directed edges, yielding $d=52$ and a flattened target dimension of $H\times d=832$. Compared with Experiment~1, this regime introduces multiple service outputs, richer dependency caps, cross-layer coupling, and more active temporal and cascading constraints. It therefore serves as the primary test of whether scenario plausibility and structural admissibility diverge as dependency complexity increases. Explicit edge sets and generator settings are reported in Appendix~\ref{app:benchmark_generator_specification}, while complete regime-specific settings are provided in the Supplementary Material.

\FloatBarrier

\subsection{Scenario-Management Strategies and Evaluation}
\label{subsec:methods_dataset_metrics}

For each regime, a controlled stochastic simulator generates 3000 reference trajectories of length 80 by evolving node and allowed-edge states under exogenous degradations, recovery dynamics, and cascading dependencies. These trajectories constitute the data from which the conditional continuation model is learned. The simulator is intentionally mostly, but not perfectly, constraint-consistent, avoiding both a trivially admissible benchmark and one dominated by structural inconsistency.

Observation windows of length $W=8$ may include noisy node states, noisy edge states, and observation masks. From each observation window, the learned model generates future scenario continuations of horizon $H=16$. Constraint rules, admissibility indicators, and violation scores are withheld from the generator and used only during post-generation validation. The data are split into training, validation, and test subsets using fractions of 0.70, 0.15, and 0.15, and results are aggregated across seeds 2026, 2027, and 2028.

The primary comparison considers unconstrained generation, hard filtering, and soft weighting. Optional projection is additionally reported as an illustrative deterministic repair baseline. The compared treatments are summarized in Table~\ref{tab:strategies}. For soft weighting, the penalty parameter is selected from
\begin{equation}
\lambda\in\{1,2,5,10,20,50,100\}.
\label{eq:lambda_sweep}
\end{equation}
The penalty parameter is selected using validation contexts. Candidate values with an effective sample size ratio of at least 0.20 are ranked first by increasing reference-simulator probability mean squared error, then by increasing weighted invalid probability mass, with the smaller $\lambda$ used as a final tie-breaker. The selected value is $\lambda=100$ for all three seeds in both regimes. No test-context information is used during selection; complete sensitivity results are reported in the Supplementary Material.

\begin{table}[htbp]
\caption{Compared scenario-management strategies. Optional projection is included as an optional deterministic repair baseline.}
\label{tab:strategies}
\centering
\setlength{\tabcolsep}{4pt}
\begin{tabularx}{\textwidth}{@{}>{\raggedright\arraybackslash}Xccc>{\raggedright\arraybackslash}X@{}}
\toprule
\textbf{Strategy} &
\shortstack{\textbf{Modifies}\\\textbf{scenarios}} &
\shortstack{\textbf{Modifies}\\\textbf{weights}} &
\shortstack{\textbf{Rejects}\\\textbf{scenarios}} &
\textbf{Role} \\
\midrule
Unconstrained diffusion & No & No & No & Learned continuation baseline \\
Hard filtering & No & No & Yes & Restriction to admissible scenarios \\
Soft weighting & No & Yes & No & Conservative scenario reweighting \\
Optional projection & Yes & No & No & Deterministic scenario repair \\
\bottomrule
\end{tabularx}
\end{table}

The evaluation distinguishes scenario admissibility from ensemble usability. Structural admissibility is measured through trajectory-level and step-level constraint-violation rates, invalid probability mass, and violation severity. The trajectory-level violation rate is
\begin{equation}
\mathrm{CVR}_{\mathrm{traj}}
=
\frac{1}{N}
\sum_{i=1}^{N}
\mathbf{1}\{I_C(\tau_i)=0\},
\label{eq:trajectory_cvr}
\end{equation}
and the step-level violation rate is
\begin{equation}
\mathrm{CVR}_{\mathrm{step}}
=
\frac{1}{NH}
\sum_{i=1}^{N}\sum_{h=1}^{H}
\mathbf{1}\{I_C(\tau_{i,h})=0\}.
\label{eq:step_cvr}
\end{equation}
For unweighted ensembles, invalid probability mass is
\begin{equation}
\widehat{p}_{\mathrm{invalid}}
=
\frac{1}{M}
\sum_{m=1}^{M}
\mathbf{1}\{I_C(\tau^{(m)})=0\}.
\label{eq:invalid_mass}
\end{equation}
When computed over the same pooled generated trajectories, \cref{eq:trajectory_cvr,eq:invalid_mass} coincide. For soft-weighted ensembles,
\begin{equation}
\widehat{p}^{\mathrm{soft}}_{\mathrm{invalid}}
=
\sum_{m=1}^{M}w^{(m)}\mathbf{1}\{I_C(\tau^{(m)})=0\}.
\label{eq:weighted_invalid_mass}
\end{equation}
Violation severity is summarized through the continuous score $\phi_C(\tau)$, both globally and by constraint family.

For hard filtering, sample efficiency is measured by the retained ratio
\begin{equation}
r_{\mathrm{hard}}=
\frac{|\mathcal S_{\mathrm{hard}}(O_t,C)|}{|\mathcal S_{\theta}(O_t)|}.
\label{eq:retained_ratio}
\end{equation}
The zero-retained-context rate is monitored and reported when nonzero or diagnostically relevant. For soft weighting, sample efficiency is measured through
\begin{equation}
\mathrm{ESS}=\frac{1}{\sum_{m=1}^{M}(w^{(m)})^2},
\label{eq:ess}
\end{equation}
and
\begin{equation}
\mathrm{ESS}_{\mathrm{ratio}}=\frac{\mathrm{ESS}}{M}.
\label{eq:ess_ratio}
\end{equation}
When reported for hard filtering, $\mathrm{ESS}_{\mathrm{hard}}/M=|\mathcal S_{\mathrm{hard}}|/M$ is computed under uniform weights over retained scenarios.

Retention and ESS quantify how much of the generated scenario ensemble remains usable after structural validation. Diversity measures whether scenario management suppresses the variability of alternative continuations. The primary diversity measure is computed context-wise as the mean pairwise Euclidean distance between flattened future trajectories and is then aggregated over test contexts and seeds. For soft weighting, the corresponding weight-sensitive diversity is retained as a separate diagnostic; the primary comparison continues to report the unweighted sample diversity. Full metric details are provided in the Supplementary Material.

Probability calibration is evaluated as a secondary ensemble-quality property relative to the reference simulator. For each test context $i$, the hidden current simulator state initializes $R=256$ independent reference-simulator rollouts over the same prediction horizon. A rollout is considered positive if, at some future step, the regime-specific service variable reaches the high-service threshold of 0.70 ($K$ in the compact regime and $K_2$ in the medium-complexity regime) and the continuation prefix through that step is structurally admissible. Let $Y_{ir}^{\mathrm{ref}}\in\{0,1\}$ denote this event for reference rollout $r$. The corresponding reference probability is
\begin{equation}
q_i=
\frac{1}{R}
\sum_{r=1}^{R}
Y_{ir}^{\mathrm{ref}},
\qquad R=256.
\label{eq:reference_probability}
\end{equation}
The generated scenario ensemble provides an analogous probability estimate $\hat p_i$ for the same event.

Agreement between the generated and reference-simulator probabilities is measured through the reference-simulator probability mean squared error
\begin{equation}
\mathrm{RS\mbox{-}MSE}
=
\frac{1}{N}
\sum_{i=1}^{N}
(\hat p_i-q_i)^2.
\label{eq:brier_score}
\end{equation}
Probability calibration is further summarized through a reference-simulator Expected Calibration Error (RS-ECE), using the standard ECE binning construction \cite{guo2017calibration}:
\begin{equation}
\mathrm{RS\mbox{-}ECE}
=
\sum_{b=1}^{B}
\frac{|B_b|}{N}
\left|
\overline{\hat p}_{B_b}
-
\overline{q}_{B_b}
\right|,
\label{eq:ece}
\end{equation}
where $B=10$ equally spaced bins are defined over the generated probability estimates in $[0,1]$, $\overline{\hat p}_{B_b}$ and $\overline{q}_{B_b}$ denote the mean generated and reference probabilities within bin $B_b$, respectively, and empty bins do not contribute. Exact event predicates and calibration details are reported in the Supplementary Material.

Robustness is evaluated without retraining under increasing observation noise and missingness, and under topology perturbations that progressively reduce the availability of allowed edges. Perturbation magnitudes, observation-probability decrements, and edge-availability degradation settings are fixed before evaluation and are applied consistently across the compared scenario-management strategies. These tests assess whether learned scenario continuations and their structural validation remain stable when the simulation context differs from the nominal condition. They do not imply that post-generation validation resolves distribution shift, which remains a distinct uncertainty problem \cite{ovadia2019trust,abdar2021review}. Full perturbation settings are reported in the Supplementary Material.

\subsection{Reproducibility}
\label{subsec:methods_reproducibility}

All simulation, training, continuation, validation, and evaluation results are aggregated over three random seeds: 2026, 2027, and 2028. The Supplementary Material reports the experimental specifications, simulator and constraint configurations, metric definitions, calibration protocol, robustness settings, and supporting seed-level results required to document the experiments reported in this study. The complete reproducibility package, including source code, machine-readable configuration files, run manifests, and processed experimental artifacts, is available from the corresponding author upon reasonable request.

\section{Results}
\label{sec:results}

\subsection{Cross-Regime Scenario-Continuation Results}
\label{subsec:results_cross_regime}

All results are aggregated over three random seeds (2026--2028), using 450 test contexts and 512 generated scenario continuations per context. Reference-simulator sanity diagnostics are computed over all 3000 sampled $H$=16 future targets per seed.

The reference simulator produces a trajectory-level constraint-violation rate of 0.030222 in the compact regime and 0.050667 in the medium-complexity regime. These values confirm that the benchmark is neither trivially admissible nor dominated by simulator-level structural inconsistency.

Table~\ref{tab:cross_experiment_results} compares the learned continuations across the two simulation regimes. With the same continuation architecture and training protocol, unconstrained invalid probability mass increases from 0.002996 in the compact regime to 0.155929 in the medium-complexity regime. Correspondingly, the fraction of scenarios retained by hard validation decreases from 0.997004 to 0.844071.

\begin{figure}[tp]
\centering
\includegraphics[width=0.84\textwidth]{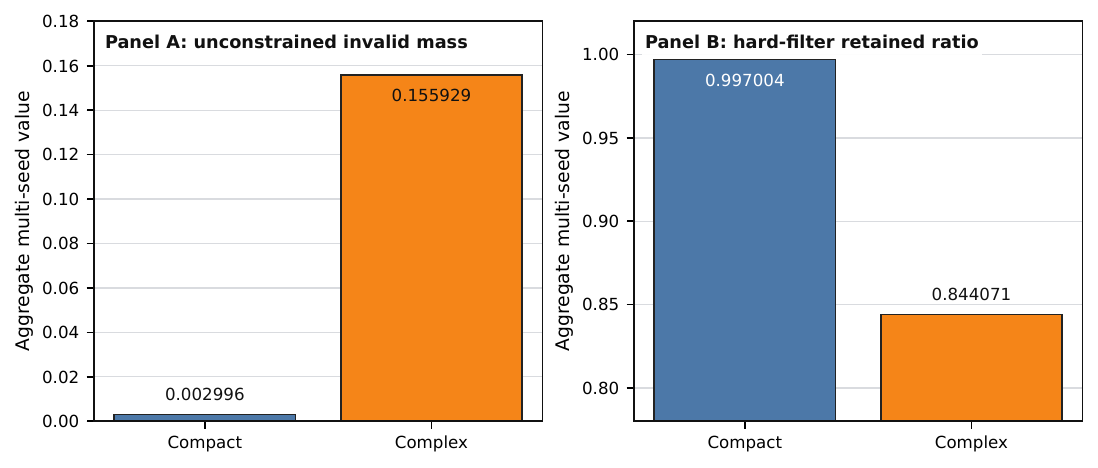}
\caption{Cross-regime comparison of learned scenario continuation. Panel A reports the invalid probability mass assigned by the unconstrained continuation model; Panel B reports the fraction of generated scenarios retained after hard structural validation.}
\label{fig:cross_experiment_structural_effect}
\end{figure}

The contrast is visualized in Figure~\ref{fig:cross_experiment_structural_effect}. In the compact positive-control regime, nearly all learned continuations satisfy the structural conditions of the simulation model. In the medium-complexity regime, the generator continues to produce a heterogeneous ensemble, but a non-negligible fraction of the generated scenarios violates the explicit admissibility conditions.

These results provide the main answer to the first research question. The medium-complexity regime exhibits a substantially larger divergence between the learned continuation distribution and the admissible scenario space. This difference is observed with the generator architecture and training protocol held fixed across regimes and is associated with the richer dependency structure of the second simulation model.

Soft weighting preserves effective sample size in both regimes but only modestly reduces invalid probability mass. The consequences of the different scenario-management strategies are examined in the following subsection.

\begin{table}[htbp]
\caption{Cross-regime comparison of learned scenario continuation and structural validation. The medium-complexity regime exhibits a larger divergence between the generated scenario distribution and the structurally admissible scenario space.}
\label{tab:cross_experiment_results}
\centering
\small
\setlength{\tabcolsep}{4pt}
\renewcommand{\arraystretch}{1.08}
\begin{tabularx}{\textwidth}{@{}>{\raggedright\arraybackslash}p{0.27\textwidth}
>{\centering\arraybackslash}p{0.13\textwidth}
>{\centering\arraybackslash}p{0.13\textwidth}
>{\raggedright\arraybackslash}X@{}}
\toprule
\textbf{Metric} &
\shortstack{\textbf{Experiment 1}\\\textbf{compact}} &
\shortstack{\textbf{Experiment 2}\\\textbf{complex}} &
\textbf{Interpretation} \\
\midrule
Nodes / edges & 9 / 16 & 15 / 37 & Dependency structure becomes more complex. \\
State dimension / target dimension & 25 / 400 & 52 / 832 & Scenario-continuation dimensionality increases. \\
Simulator trajectory CVR & 0.030222 & 0.050667 & Both reference regimes remain within the intended validity range. \\
Unconstrained invalid mass & 0.002996 & 0.155929 & Learned continuations increasingly leave the admissible scenario space. \\
Hard retained ratio & 0.997004 & 0.844071 & Structural validation becomes consequential in the complex regime. \\
Soft invalid mass & 0.002715 & 0.148807 & Reweighting only modestly reduces invalid scenario mass. \\
Soft ESS ratio & 0.999930 & 0.998764 & Reweighting preserves the effective ensemble size. \\
Unconstrained diversity & 1.030889 & 1.561000 & The complex regime produces a more dispersed continuation ensemble. \\
\bottomrule
\end{tabularx}
\end{table}

\FloatBarrier

\subsection{Structural Validation in the Complex Regime}
\label{subsec:results_constraint_handling}

The medium-complexity regime provides the primary test of post-generation scenario validation because its unconstrained continuation ensemble assigns non-negligible probability mass to structurally inadmissible scenarios. Table~\ref{tab:experiment_2_strategy_results} compares hard filtering and soft weighting against unconstrained generation and additionally reports optional projection as a deterministic repair baseline.

Hard filtering removes all invalid scenarios from the retained ensemble. Invalid mass decreases from 0.155929 to zero, while the retained ratio is 0.844071. Thus, approximately 84.4\% of the generated continuations remain available for subsequent simulation analysis after enforcing all structural conditions. In this regime, hard filtering provides an explicit admissibility guarantee without collapsing the usable scenario ensemble.

Soft weighting preserves all generated scenarios and maintains an ESS ratio of 0.998764. However, weighted invalid mass decreases only from 0.155929 to 0.148807, an absolute reduction of 0.007122. Soft weighting therefore preserves ensemble size and continuity, but does not materially change the Boolean admissibility of the underlying scenarios. Its effect is primarily a redistribution of probability mass within the generated ensemble.

For comparison, optional projection reduces invalid mass to 0.000023 while retaining all generated scenarios. This result is obtained through the fixed single-pass deterministic repair procedure described in Section~\ref{subsec:methods_symbolic_constraints}, rather than through scenario selection or probability reweighting. The repaired ensemble therefore no longer coincides with the original learned continuation distribution. Because repair distance and distributional distortion are not evaluated in the present study, the near-complete admissibility of this baseline should not be interpreted as evidence of methodological superiority over hard filtering or soft weighting.

The primary comparison therefore exposes a trade-off between two distinct scenario-management choices. Hard filtering guarantees admissibility at the cost of scenario rejection, whereas soft weighting preserves the original scenarios but offers limited admissibility improvement. Optional projection separately shows that direct deterministic intervention can preserve ensemble cardinality while substantially reducing violations, at the cost of altering scenario content.

\begin{table}[htbp]
\caption{Scenario-management results in the medium-complexity simulation regime. Hard filtering and soft weighting constitute the primary comparison; optional projection is reported separately as an illustrative deterministic repair baseline.}
\label{tab:experiment_2_strategy_results}
\centering
\footnotesize
\setlength{\tabcolsep}{5pt}
\renewcommand{\arraystretch}{1.08}
\begin{tabularx}{\textwidth}{@{}>{\raggedright\arraybackslash}p{0.22\textwidth}
*{6}{>{\centering\arraybackslash}p{0.105\textwidth}}@{}}
\toprule
\textbf{Strategy} &
\shortstack{\textbf{Invalid}\\\textbf{mass}} &
\shortstack{\textbf{Trajectory}\\\textbf{CVR}} &
\shortstack{\textbf{Retained}\\\textbf{ratio}} &
\shortstack{\textbf{ESS}\\\textbf{ratio}} &
\textbf{Diversity} &
\textbf{RS-ECE} \\
\midrule
Unconstrained diffusion & 0.155929 & 0.155929 & 1.000000 & 1.000000 & 1.561000 & 0.070736 \\
Hard filtering & 0.000000 & 0.000000 & 0.844071 & 0.844071 & 1.539848 & 0.070643 \\
Soft weighting & 0.148807 & 0.155929 & 1.000000 & 0.998764 & 1.561000 & 0.070731 \\
Optional projection & 0.000023 & 0.000023 & 1.000000 & 1.000000 & 1.560327 & 0.070284 \\
\bottomrule
\end{tabularx}
\end{table}

\FloatBarrier

\subsection{Sources of Scenario Inadmissibility}
\label{subsec:results_family_diagnostics}

Table~\ref{tab:family_diagnostics} decomposes the invalid probability mass of the medium-complexity continuation ensemble by constraint family.

Dependency constraints account for nearly all observed structural inadmissibility, with an invalid mass of 0.155929. Threshold and temporal-ordering violations remain limited to 0.000370 and 0.000383, respectively, while mutual-exclusion and structural-invariant violations make no aggregate contribution.

The learned continuation model therefore preserves basic state ranges, topology-related invariants, maintenance exclusions, and most temporal prerequisites. The main difficulty lies instead in maintaining quantitative consistency between downstream variables and their multiple upstream dependencies.

This distinction is important when validating learned continuations against the structural semantics of the simulation model. Individual state trajectories may remain numerically plausible and locally well-behaved while their joint evolution violates the dependency structure of the simulated system. Aggregate or marginal trajectory diagnostics would not necessarily expose this failure mode; explicit relational validation is required.

Family-level invalid masses are not additive because a generated scenario may violate more than one constraint family simultaneously.

\begin{table}[ht]
\caption{Sources of structural inadmissibility in the medium-complexity scenario-continuation ensemble. Dependency-consistency violations dominate, whereas threshold, temporal-ordering, mutual-exclusion, and structural-invariant violations remain negligible.}
\label{tab:family_diagnostics}
\centering
\begin{tabularx}{\textwidth}{lcX}
\toprule
\textbf{Constraint family} & \textbf{Invalid mass} & \textbf{Interpretation} \\
\midrule
Dependency & 0.155929 & Dominant mismatch between learned continuations and simulation dependencies. \\
Threshold & 0.000370 & Minimal prerequisite-threshold mismatch. \\
Temporal ordering & 0.000383 & Minimal temporal-consistency mismatch. \\
Mutual exclusion & 0.000000 & No aggregate incompatibility contribution. \\
Structural invariant & 0.000000 & No aggregate graph-state-invariant contribution. \\
\bottomrule
\end{tabularx}
\end{table}

\subsection{Scenario-Level Probability Calibration}
\label{subsec:results_calibration}

Probability calibration is evaluated as a secondary property of the generated scenario ensemble. For each test context, the ensemble is used to estimate the probability of a regime-specific admissible high-service event.

Across both regimes, scenario-management strategies have only a marginal effect on the reference-simulator probability mean squared error (RS-MSE) and reference-simulator Expected Calibration Error (RS-ECE). In the compact regime, hard filtering changes the RS-MSE from 0.578380 to 0.577716 and the RS-ECE from 0.741424 to 0.740841. In the medium-complexity regime, RS-ECE changes from 0.070736 to 0.070643 after hard filtering.

These results show that structural validation and probability calibration capture different properties of an ensemble. Removing structurally inadmissible scenarios does not necessarily improve the correspondence between estimated event probabilities and the corresponding probabilities induced by the reference simulator.

Calibration values should also not be compared directly across the two regimes as measures of simulation difficulty. The compact and medium-complexity experiments use regime-specific events with different operating ranges and reference-simulator probabilities. The calibration results are therefore interpreted within each regime and across scenario-management strategies, rather than as a cross-regime ranking.

\subsection{Robustness of Scenario Continuation and Validation}
\label{subsec:results_robustness}

Robustness is evaluated without retraining under perturbed simulation contexts. Observation perturbations increase noise and missingness in the scenario history, whereas topology perturbations progressively reduce the availability of selected allowed dependencies.

In the medium-complexity regime, severe observation perturbation increases invalid mass from 0.154819 to 0.158594 and reduces the hard-filter retained ratio from 0.845181 to 0.841406. The structural-validation burden therefore increases only moderately when the observed scenario history is degraded.

Topology perturbation produces a larger effect. Under the severe condition, invalid mass rises to 0.169536, and the retained ratio falls to 0.830464. This result is consistent with the family-level analysis: because dependency consistency is the principal source of inadmissibility, degradation of allowed-edge availability affects scenario validation more strongly than observation noise alone.

The perturbation study indicates that the continuation-and-validation workflow degrades gradually under the examined conditions. It does not establish robustness to arbitrary distribution shifts, nor does post-generation validation compensate for a misspecified simulation structure. Detailed values are reported in Table~\ref{tab:app_robustness_summary}.

\subsection{Answers to the Research Questions}
\label{subsec:results_summary}

The experimental results provide direct answers to the two research questions introduced in Section~\ref{sec:introduction}.

\noindent\textbf{RQ1: Structural-complexity comparison.}
With the continuation architecture, training protocol, prediction horizon, ensemble size, and evaluation procedure held fixed, unconstrained invalid mass increases from 0.002996 in the compact regime to 0.155929 in the medium-complexity regime. The medium-complexity regime therefore exhibits a substantially larger divergence between the learned continuation distribution and the admissible scenario space. This divergence is dominated by multi-parent dependency constraints rather than by basic state-range or topology violations.

\noindent\textbf{RQ2: Trade-offs among scenario-management strategies.}
Hard filtering provides the strongest admissibility guarantee and retains 84.4\% of the generated scenarios in the complex regime. Soft weighting preserves an ESS ratio of 0.998764 but reduces invalid mass only modestly. The two strategies therefore address different simulation requirements: validated scenario selection and sample-preserving probability reweighting. Optional projection provides an ancillary comparison: direct deterministic modification reduces invalid mass to 0.000023 while preserving ensemble cardinality, but changes the generated trajectories and is not part of the primary RQ2 comparison.

Across both questions, probability calibration changes only marginally under structural validation. Scenario admissibility, ensemble usability, and event-probability calibration should consequently be treated as separate evaluation dimensions.

\FloatBarrier

\FloatBarrier
\section{Discussion}
\label{sec:discussion}

\subsection{Learned Scenario Continuation under Structural Complexity}
\label{subsec:discussion_main}

The experiments show a substantial difference in the admissibility of learned scenario continuations across the two structural-complexity regimes. In the compact positive-control regime, the conditional diffusion model generates an ensemble that is already almost entirely contained within the admissible scenario space. Under these conditions, the external validation layer acts primarily as a certification mechanism: it confirms the structural consistency of the generated continuations while rejecting only a negligible fraction of the ensemble.

The same continuation architecture behaves differently in the medium-complexity regime. With the continuation architecture and training protocol held fixed, a non-negligible fraction of the learned continuations violates the structural conditions of the simulation model. The difference is therefore not characterized solely by the increased state dimension. The second regime also introduces multi-parent dependencies, hierarchical outputs, cross-layer coupling, and temporal prerequisites that must remain jointly consistent throughout the generated scenario.

This result provides the central simulation-oriented finding of the study. A learned continuation model may reproduce local trajectory patterns while failing to preserve the relational structure that defines allowable system evolution. In the structurally richer regime examined here, scenario generation and scenario validation can no longer be treated as equivalent tasks. The continuation model determines which futures are statistically supported by the learned distribution, whereas the structural layer determines which of those futures are admissible under the explicit simulation model.

\subsection{Plausibility and Admissibility as Distinct Scenario Properties}
\label{subsec:discussion_reliability}

The distinction between statistical plausibility and structural admissibility is not merely terminological. It identifies two different sources of evidence about a generated scenario. Statistical plausibility is inherited from the learned conditional distribution: a scenario is plausible to the extent that it is supported by the trajectories used to train the continuation model. In the present study, this term denotes support under the learned conditional distribution and is not intended as an independent measure of distributional fidelity to the reference simulator. Structural admissibility is established against explicit domain relations, prerequisites, temporal conditions, and graph-state invariants.

The experiments demonstrate that these properties can diverge. In the complex regime, the continuation model assigns non-negligible probability mass to scenarios that violate the symbolic representation of the simulated system. These scenarios are not necessarily characterized by implausible individual values. Instead, their joint evolution is inconsistent with dependencies that are evaluated only after generation.

For data-driven simulation, this separation has an important methodological consequence. Evaluation based only on predictive error, likelihood, or marginal trajectory statistics may fail to identify scenarios whose components are individually well behaved but jointly inconsistent. Conversely, structural validation cannot establish that a scenario is well represented by the learned distribution or that its estimated probability is calibrated. Scenario plausibility, structural admissibility, and probability calibration should therefore be treated as complementary rather than hierarchical properties.

This layered interpretation also clarifies the scope of the proposed method. The symbolic layer validates generated continuations relative to the specified constraint set $C$; it does not establish that $C$ is complete, that the reference simulator is an accurate representation of a real system, or that the learned continuation distribution generalizes beyond the evaluated conditions. The method supports scenario validation within a simulation workflow, not validation of the entire simulation study against reality.

\subsection{Trade-offs among Scenario-Management Strategies}
\label{subsec:discussion_constraint_strategies}

Hard filtering and soft weighting address two distinct requirements of post-generation scenario management, while optional projection provides a separate illustrative repair baseline.

Hard filtering provides the clearest admissibility semantics. Every retained continuation satisfies the implemented structural conditions. In the medium-complexity regime, this guarantee is obtained while retaining 84.4\% of the generated ensemble. Hard filtering is therefore appropriate when downstream simulation analysis must operate exclusively on valid scenarios and when rejection of part of the ensemble is acceptable. Its limitation is direct: as invalid probability mass increases, the number of usable scenarios decreases, and some observation contexts may eventually require additional sampling.

Soft weighting preserves the complete generated ensemble and almost all of its effective sample size, with an ESS ratio of 0.998764 in the medium-complexity regime. This behavior is useful when preservation of the generated sample set and scenario coverage are more important than Boolean admissibility. However, the limited reduction in weighted invalid mass shows that reweighting should not be interpreted as structural enforcement. The scenarios themselves remain unchanged; only their contribution to ensemble-level estimates is modified. Soft weighting is therefore better understood as a graded preference mechanism than as a validation guarantee.

Optional projection occupies a different methodological position. Its fixed single-pass deterministic transformation preserves ensemble cardinality and achieves near-complete admissibility by modifying the generated trajectories. The repaired ensemble is therefore no longer sampled directly from the original learned continuation distribution. Moreover, because the present evaluation does not quantify repair distance or distributional distortion, this result demonstrates the effect of direct rule-based intervention rather than the superiority of repair as a scenario-management method.

These observations suggest that the primary scenario-management choice depends on whether a simulation workflow prioritizes admissibility of retained scenarios or preservation of the generated sample set under probability reweighting. Optional projection represents a separate intervention-oriented baseline whose fuller assessment would additionally require modification-related diagnostics such as repair distance and distributional distortion.

\subsection{Dependency Preservation as the Main Modeling Challenge}
\label{subsec:discussion_dependency_failures}

The family-level analysis identifies dependency consistency as the dominant source of scenario inadmissibility. Threshold, temporal-ordering, mutual-exclusion, and structural-invariant violations remain negligible, whereas dependency violations account for nearly all invalid probability mass in the medium-complexity regime.

This pattern indicates that the continuation model learns basic state ranges and most local regularities more readily than the quantitative relations connecting multiple upstream and downstream variables. A generated scenario may therefore remain smooth, bounded, and locally plausible while violating the multivariate dependencies encoded by the simulation model. This failure mode would be difficult to detect through inspection of individual state variables or aggregate trajectory statistics alone.

From a modeling perspective, the result motivates explicit relational diagnostics whenever learned components are used to extend graph-structured simulations. It also suggests a direction for future model development. Constraint-guided generation, graph-aware denoisers, relational regularization, or hybrid simulation architectures may reduce the dependency mismatch before post-generation validation is applied. The present experiments do not compare these alternatives, but they establish a controlled baseline against which such methods can be evaluated.

\subsection{Implications for Data-Driven Simulation and Scenario Management}
\label{subsec:discussion_generative_decision_support}

The proposed flow-map formulation positions conditional diffusion as a learned scenario-continuation component within a broader simulation workflow. The reference simulator generates trajectories that define the training environment; the learned model extends partial histories into alternative futures; and the symbolic layer validates and manages the resulting ensemble. This division of responsibilities avoids treating the learned generator as a replacement for the underlying simulator.

The framework is particularly relevant when simulation analysis requires many alternative continuations from a common intermediate state. Repeated execution of a detailed simulator may be costly, difficult to configure, or unsuitable for rapid ensemble exploration. A learned continuation model can provide an additional mechanism for generating candidate futures, while structural validation prevents the learned component from being evaluated solely through its own statistical support.

The results also have implications for scenario management. A generated ensemble should not be regarded as a homogeneous collection of interchangeable futures. Scenarios may differ in admissibility, violation severity, probability weight, and sensitivity to observation or topology perturbations. Retaining these distinctions enables a simulation workflow to support different downstream uses, including admissible-scenario selection, weighted ensemble analysis, stress testing, and repair-oriented exploration.

At the same time, the experiments support a cautious interpretation of learned simulation components. Structural validation can identify violations of known constraints, but it cannot recover omitted domain knowledge or correct a misspecified reference model. The usefulness of the validated ensemble therefore depends on both the learned continuation model and the quality of the explicit simulation semantics against which it is checked.

\subsection{Robustness and Admissibility Boundaries}
\label{subsec:discussion_robustness}

The perturbation experiments show a gradual degradation of the continuation-and-validation workflow under the examined observation and topology changes. Increased observation noise and missingness produce only a moderate increase in invalid mass, whereas topology perturbations have a larger effect. This difference is consistent with the constraint-family analysis, because dependency relations constitute the principal source of inadmissibility in the complex regime.

These results should be interpreted as diagnostics within the tested perturbation range, not as a general robustness guarantee. Observation perturbations alter the information available to the continuation model, while topology perturbations degrade the availability of selected allowed dependencies and thereby alter the structural context in which generated scenarios are evaluated. More severe or qualitatively different shifts could affect both the learned distribution and the relevance of the constraint set.

The validation layer also has a clear boundary: it can detect scenarios that violate the encoded model, but it does not resolve distribution shift or establish external admissibility. If the graph topology, constraint set, or event definitions change in operation, the simulation workflow may require constraint revision, generator adaptation, or retraining. Robust scenario generation and structural validation should therefore be treated as related but distinct concerns.

\subsection{Limitations and Future Work}
\label{subsec:discussion_limitations}

Several limitations define the scope of the present study. First, the benchmark is synthetic and is intended as a controlled methodological testbed rather than as an operational validation. This design enables a controlled comparison across graph-complexity regimes and post-generation strategies, but it does not reproduce the full uncertainty, heterogeneity, or semantic ambiguity of a real simulation domain.

Moreover, only two graph regimes are considered. They differ simultaneously in dimensionality, dependency density, hierarchy, temporal structure, and selected regime-specific parameters. The experiments therefore establish a substantial cross-regime difference associated with structural complexity but do not disentangle the contribution of each structural factor. A broader experimental design should vary graph size, edge density, dependency order, temporal depth, constraint overlap, and sampling parameters independently.

Third, the structural conditions are manually specified and treated as correct and complete. In practical simulation studies, constraints may be uncertain, partially observed, inconsistent, or time-varying. Future work should investigate robust validation under uncertain constraints, sensitivity to constraint misspecification, and mechanisms for revising the admissibility model as simulation knowledge evolves.

Fourth, the generator is intentionally prevented from accessing the symbolic constraints during training. This isolates post-generation validation, but does not compare it with constraint-aware training, guided diffusion, graph-structured denoisers, or hybrid simulation models. Such comparisons would clarify when inadmissibility is best addressed during learning, during sampling, or after scenario generation.

Fifth, the present evaluation concentrates on structural admissibility and ensemble usability. Additional simulation-oriented measures should assess distributional fidelity to held-out simulator trajectories, temporal event statistics, coverage of rare but admissible scenarios, repair distance, computational cost, and the effect of scenario management on downstream simulation conclusions.

Future work should therefore extend the benchmark to larger and more heterogeneous simulation models, introduce controlled ablations of structural complexity, compare post-generation validation with constraint-integrated generation, and evaluate the method on operational or high-fidelity simulation environments. A further direction is the development of adaptive workflows in which the generator, simulator, and validation layer exchange information while preserving a transparent distinction between learned statistical behavior and explicit simulation semantics.

\section{Conclusions}
\label{sec:conclusions}

This paper presented a method for learned scenario continuation and post-generation structural validation in dynamic graph simulations. A conditional diffusion model extends partial simulation histories into ensembles of alternative future trajectories, while an external symbolic layer evaluates whether the generated continuations satisfy the structural conditions of the simulated system. Hard filtering and soft weighting provide two primary mechanisms for managing the resulting scenario ensemble, while optional projection is reported separately as an illustrative deterministic repair baseline.

The controlled experiments reveal a substantial cross-regime difference associated with structural complexity. In the compact positive-control regime, the learned continuation distribution is almost entirely contained within the admissible scenario space. With the continuation architecture and training protocol held fixed, invalid probability mass increases substantially in the medium-complexity regime. The divergence is driven primarily by multi-parent dependency constraints rather than by state-range, mutual-exclusion, or basic temporal violations. These results show that learning locally plausible trajectories does not necessarily preserve the relational structure required by a simulation model.

The comparison of the two primary scenario-management strategies exposes complementary trade-offs. Hard filtering guarantees admissibility for the retained ensemble and preserves 84.4\% of the generated scenarios in the complex regime. Soft weighting retains the complete ensemble and an ESS ratio of 0.998764, but only modestly reduces invalid probability mass. Optional projection additionally shows that direct deterministic modification can achieve near-complete admissibility while preserving ensemble cardinality, but the repaired scenarios no longer coincide with the original generated trajectories. Its result is therefore interpreted as an illustrative repair baseline rather than as evidence of superiority over the two primary strategies.

More broadly, the study establishes statistical plausibility and structural admissibility as distinct properties of learned simulation scenarios. Probability calibration constitutes a further, complementary property: removing inadmissible scenarios does not automatically improve event-probability calibration. Evaluation of data-driven simulation components should therefore distinguish the quality of the learned continuation distribution, compliance with explicit simulation semantics, and usability of the resulting ensemble.

The present benchmark is intentionally synthetic and does not establish the external validity of the reference simulator or the completeness of its constraint set. Nevertheless, it provides a controlled basis for studying learned scenario continuation across different levels of structural complexity and for comparing post-generation validation strategies. Future work should extend the analysis to larger and more heterogeneous simulation models, isolate individual dimensions of structural complexity, compare post-generation validation with constraint-guided generation, and evaluate distributional fidelity, rare-scenario coverage, repair distance, and computational cost in operational or high-fidelity simulation environments.

Overall, the results indicate that learned scenario generation should not be assessed through statistical support alone. As simulation models become structurally richer, explicit validation of generated continuations becomes a necessary component of reliable scenario generation and management.

\section*{Author Contributions}

Conceptualization: M.R.d.S. and G.L.P.;
methodology: M.R.d.S. and G.L.P.;
software: M.R.d.S.;
validation: M.R.d.S. and A.M.;
formal analysis: M.R.d.S. and G.L.P.;
investigation: M.R.d.S.;
writing---original draft: M.R.d.S.;
writing---review and editing: M.R.d.S. and A.M.;
supervision: A.M.

\section*{Funding}
This research received no external funding.

\section*{Institutional Review Board Statement}
Not applicable.

\section*{Informed Consent Statement}
Not applicable.

\section*{Data Availability Statement}
The Supplementary Material provides the experimental specifications, parameter settings, metric definitions, calibration and robustness protocols, and supporting seed-level results required to document the findings reported in this study. The complete reproducibility package, including source code, machine-readable configuration files, run manifests, and processed experimental artifacts, is available from the corresponding author upon reasonable request.

\section*{Acknowledgments}
The authors gratefully acknowledge the ``HPC4AI'' initiative~\cite{aldinucci_hpc4ai_2018} for providing access to high-performance computing resources that supported preliminary implementation of this work.

\section*{Declaration of Competing Interest}
The authors declare no conflicts of interest.

\appendix
\numberwithin{equation}{section}
\numberwithin{figure}{section}
\numberwithin{table}{section}

\section{Benchmark Specification} 
\label{app:benchmark_generator_specification}

This appendix documents the benchmark specification supporting the scenario-continuation and structural-validation experiments presented in Section~\ref{sec:results}.

The main paper compares learned scenario admissibility across two structural-complexity regimes. Accordingly, this appendix reports the graph topologies, node semantics, and shared continuation-model components needed to interpret the controlled benchmark; complete simulator and experimental specifications are provided in the Supplementary Material.

The benchmark consists of two dynamic-graph simulation regimes implemented within the same reference-simulation framework and validation pipeline. They use the same learned continuation architecture and training protocol, while retaining regime-specific simulator and sampling parameters. The regimes differ principally in graph dimensionality and dependency complexity.

\subsection{Experiment 1: Compact Graph Regime}
\label{app:experiment_1_specification}

The compact-regime node set is introduced in \cref{subsubsec:methods_experiment_1}. The allowed directed edge set used to instantiate that topology is
\[
\begin{split}
E_{\mathrm{compact}}
=
\{&
P\!\to\!C,\,
P\!\to\!D,\,
P\!\to\!S,\,
Y\!\to\!C,\,
Y\!\to\!D,\,
C\!\to\!D,\,
S\!\to\!D,\\
&
C\!\to\!K,\,
D\!\to\!K,\,
L\!\to\!K,\,
P\!\to\!K,\,
R\!\to\!P,\,
R\!\to\!C,\\
&
R\!\to\!Y,\,
R\!\to\!S,\,
M\!\to\!K
\}.
\end{split}
\]
This edge set provides the compact topology summarized in \cref{tab:benchmark_specification,fig:graph_topologies}.

\subsection{Experiment 2: Medium-Complexity Dependency Graph Regime}
\label{app:experiment_2_specification}

The medium-complexity node set is introduced in \cref{subsubsec:methods_experiment_2}. \Cref{tab:app_complex_nodes} reports the symbol-level interpretation used in the implementation, while the allowed directed edge set is listed below.

\begin{table}[htbp]
\caption{Node set for the medium-complexity dependency graph.}
\label{tab:app_complex_nodes}
\centering
\small
\setlength{\tabcolsep}{4pt}
\renewcommand{\arraystretch}{1.06}
\begin{tabularx}{\textwidth}{@{}>{\centering\arraybackslash}p{0.09\textwidth}
>{\raggedright\arraybackslash}p{0.34\textwidth}
>{\raggedright\arraybackslash}X@{}}
\toprule
\textbf{Symbol} & \textbf{Name} & \textbf{Interpretation} \\
\midrule
$P_m$ & Main power availability & Primary energy source. \\
$P_b$ & Backup power availability & Redundant or backup energy source. \\
$Y$ & Cyber integrity & Integrity of digital control and data channels. \\
$C_1$ & Internal communication reliability & Local coordination reliability. \\
$C_2$ & External communication reliability & External coordination or network reliability. \\
$S_1$ & Primary sensing coverage & Primary monitoring or sensing capability. \\
$S_2$ & Secondary sensing coverage & Redundant sensing capability. \\
$D_1$ & Data-processing integrity & First-stage data-processing reliability. \\
$D_2$ & Data-fusion integrity & Cross-source data-fusion reliability. \\
$Q$ & Data quality/control integrity & Quality-control and consistency layer. \\
$L$ & Logistics/resource readiness & Sustained resource availability. \\
$R$ & Repair/recovery capacity & Capacity to restore degraded components. \\
$M$ & Maintenance/safe mode & Emergency or maintenance mode restricting service. \\
$K_1$ & Local service feasibility & Intermediate scenario-level service outcome. \\
$K_2$ & Global service feasibility & Final scenario-level service outcome used for ensemble calibration. \\
\bottomrule
\end{tabularx}
\end{table}

The allowed directed edge set used to instantiate the medium-complexity topology is
\[
\begin{split}
E_{\mathrm{complex}}=\{&
P_m\!\to\!C_1,\,
P_m\!\to\!D_1,\,
P_m\!\to\!S_1,\,
P_m\!\to\!K_1,\,
P_m\!\to\!Q,\\
&
P_b\!\to\!C_2,\,
P_b\!\to\!D_2,\,
P_b\!\to\!S_2,\,
P_b\!\to\!K_2,\\
&
Y\!\to\!C_1,\,
Y\!\to\!C_2,\,
Y\!\to\!D_1,\,
Y\!\to\!Q,\\
&
C_1\!\to\!D_1,\,
C_1\!\to\!K_1,\,
C_1\!\to\!D_2,\\
&
C_2\!\to\!D_2,\,
C_2\!\to\!K_2,\\
&
S_1\!\to\!D_1,\,
S_1\!\to\!Q,\,
S_2\!\to\!D_2,\,
S_2\!\to\!Q,\\
&
D_1\!\to\!D_2,\,
D_1\!\to\!K_1,\,
D_2\!\to\!K_2,\\
&
Q\!\to\!K_1,\,
Q\!\to\!K_2,\,
L\!\to\!K_1,\,
L\!\to\!K_2,\\
&
R\!\to\!P_m,\,
R\!\to\!P_b,\,
R\!\to\!Y,\,
R\!\to\!C_1,\,
R\!\to\!S_1,\,
R\!\to\!S_2,\\
&
M\!\to\!K_1,\,
M\!\to\!K_2
\}.
\end{split}
\]
This edge set provides the medium-complexity topology summarized in \cref{tab:benchmark_specification,fig:graph_topologies}.

\FloatBarrier

\subsection{Reference-Simulator Dynamics}
\label{app:reference_simulator_dynamics}

The reference simulator evolves node and allowed-edge states through stochastic degradation, recovery, and dependency-propagation updates. At trajectory or rollout initialization, the exogenous event schedule is sampled. At each simulation step, the effects of the active events are applied together with the dependency and service-variable updates according to the graph structure and regime-specific relations. The compact and medium-complexity regimes use the same overall update logic and random-seed policy, while their graph structures and dependency relations differ as specified above. Complete transition parameters, event probabilities, initialization distributions, update ordering, clipping operations, and simulator-internal plausibility controls are provided in the Supplementary Material. This separation is important because the structural validator evaluates generated continuations against an explicit constraint set, whereas the reference simulator is intentionally mostly, but not perfectly, constraint-consistent.

\subsection{Shared Diffusion Configuration}
\label{app:shared_dataset_diffusion_configuration}

Both benchmark regimes use the same conditional diffusion continuation architecture and training protocol. This controls for differences in model capacity and optimization while allowing regime-specific simulator and sampling parameters.

Dataset-level parameters are summarized in Table~\ref{tab:benchmark_specification}. Table~\ref{tab:app_diffusion_configuration} reports the shared continuation-model configuration used in both regimes. Regime-specific sampling parameters and complete training and evaluation settings are reported in the Supplementary Material.

\begin{table}[htbp]
\caption{Shared conditional diffusion scenario-continuation configuration used in both benchmark regimes.}
\label{tab:app_diffusion_configuration}
\centering
\small
\setlength{\tabcolsep}{6pt}
\renewcommand{\arraystretch}{1.05}
\begin{tabular}{@{}>{\raggedright\arraybackslash}p{0.40\textwidth}
>{\raggedright\arraybackslash}p{0.38\textwidth}@{}}
\toprule
\textbf{Component} & \textbf{Configuration} \\
\midrule
Diffusion steps during training & 100 \\
Sampling steps & 25 \\
Noise schedule & Cosine \\
Sampler & DDIM \\
Prediction type & Noise prediction \\
Denoiser backbone & Temporal TCN \\
Hidden dimension & 256 \\
Number of layers & 4 \\
Dropout & 0.10 \\
Time-embedding dimension & 64 \\
Conditioning dimension & 256 \\
Training objective & Noise-prediction MSE; see \cref{eq:diffusion_loss} \\
Structural constraints provided to continuation model & No \\
\bottomrule
\end{tabular}
\end{table}

\section{Constraint Layer Specification} \label{app:constraint_families_violation_scores}

This appendix formalizes the structural scenario-validation layer introduced in Section~\ref{subsec:methods_symbolic_constraints}.

Throughout the paper, structural admissibility is treated as a scenario property distinct from statistical plausibility and probability calibration. The objective of this appendix is to specify how the admissible scenario space is operationalized within the controlled simulation benchmark. The aggregation rules described below apply to both simulation regimes. Because the medium-complexity regime constitutes the primary structural-validation experiment, the explicit regime-specific equations in the following subsections focus on that setting; the complete compact-regime constraint specification is provided in the Supplementary Material.

Each constraint family contributes to the Boolean admissibility indicator defined in Equation~\ref{eq:validity_indicator} and to the continuous violation score defined in Equation~\ref{eq:violation_score}. These quantities support scenario filtering, reweighting, and family-level diagnostics throughout Section~\ref{sec:results}; optional projection uses the same explicit structural rules as a separate deterministic repair baseline.

Table~\ref{tab:app_constraint_families} summarizes the five constraint families and their associated violation semantics.

\begin{table}[htbp]
\caption{Constraint families used to evaluate the structural admissibility of generated scenario continuations.}
\label{tab:app_constraint_families}
\centering
\small
\setlength{\tabcolsep}{4pt}
\renewcommand{\arraystretch}{1.08}
\begin{tabularx}{\textwidth}{@{}>{\raggedright\arraybackslash}p{0.22\textwidth}
>{\raggedright\arraybackslash}X
>{\raggedright\arraybackslash}p{0.27\textwidth}@{}}
\toprule
\textbf{Constraint family} & \textbf{Role} & \textbf{Violation score} \\
\midrule
Threshold constraints &
Encode prerequisite conditions for high service or component activation. &
Threshold-margin severity. \\
Dependency constraints &
Enforce consistency between downstream variables and upstream parents. &
Excess-over-cap severity. \\
Mutual exclusion &
Prevent incompatible states from co-occurring. &
Incompatibility severity. \\
Temporal ordering &
Require selected states to recover or activate only after prerequisite stability. &
Temporal inconsistency severity. \\
Structural invariants &
Enforce bounded values, allowed-edge consistency, and basic graph-state admissibility. &
Invariant-violation severity. \\
\bottomrule
\end{tabularx}
\end{table}

For a generated scenario continuation $\tau$, each constraint family
$f \in \mathcal{F}$ produces a Boolean family indicator
$I_f(\tau)\in\{0,1\}$ and a non-negative family-level violation score
$\phi_f(\tau)\geq 0$. The scenario-level admissibility indicator introduced in
\cref{eq:validity_indicator} is implemented as the conjunction of all
family-level indicators:
\[
I_C(\tau)
=
\prod_{f \in \mathcal{F}} I_f(\tau).
\]
The continuous violation score introduced in \cref{eq:violation_score} is
implemented as a weighted aggregation of family-level margins:
\[
\phi_C(\tau)
=
\sum_{f \in \mathcal{F}} \alpha_f \phi_f(\tau),
\qquad
\alpha_f \geq 0.
\]
In the reported experiments, all five constraint families use equal weights, $\alpha_f=1$.
Boolean admissibility is used for hard filtering and family-level violation rates,
whereas the continuous score is used for severity diagnostics and soft
weighting. For upper-bound dependency constraints, violation margins are
computed as the positive excess above the admissible cap and then aggregated
over constraints and time steps.

\subsection{Threshold Constraints}
\label{app:threshold_constraints}

Threshold constraints specify prerequisite conditions for high-service outcomes
in the medium-complexity regime. For the local service variable $K_1$, high
activation requires the relevant upstream variables to satisfy their respective
threshold conditions:
\[
\begin{aligned}
K_{1,t}\geq\theta_{K_1}
\quad\Rightarrow\quad&
P_{m,t}\geq\theta_{P_m}
\land C_{1,t}\geq\theta_{C_1}
\land D_{1,t}\geq\theta_{D_1} \\
&\land Q_t\geq\theta_Q
\land L_t\geq\theta_L
\land M_t\leq\theta_M .
\end{aligned}
\]
For the global service variable $K_2$, high activation additionally requires
support from the intermediate service variable and from the global upstream
components:
\[
\begin{aligned}
K_{2,t}\geq\theta_{K_2}
\quad\Rightarrow\quad&
K_{1,t}\geq\theta_{K_1}^{\mathrm{low}}
\land P_{b,t}\geq\theta_{P_b}
\land C_{2,t}\geq\theta_{C_2} \\
&\land D_{2,t}\geq\theta_{D_2}
\land Q_t\geq\theta_Q
\land L_t\geq\theta_L
\land M_t\leq\theta_M .
\end{aligned}
\]
The threshold values used for these conditions are reported in
\cref{tab:app_threshold_values}.

\begin{table}[htbp]
\caption{Threshold values for the medium-complexity dependency graph.}
\label{tab:app_threshold_values}
\centering
\small
\setlength{\tabcolsep}{6pt}
\renewcommand{\arraystretch}{1.05}
\begin{tabular}{@{}lclc@{}}
\toprule
\textbf{Threshold} & \textbf{Value} & \textbf{Threshold} & \textbf{Value} \\
\midrule
$\theta_{P_m}$ & 0.55 & $\theta_{P_b}$ & 0.50 \\
$\theta_Y$ & 0.50 & $\theta_{C_1}$ & 0.60 \\
$\theta_{C_2}$ & 0.60 & $\theta_{S_1}$ & 0.55 \\
$\theta_{S_2}$ & 0.50 & $\theta_{D_1}$ & 0.55 \\
$\theta_{D_2}$ & 0.60 & $\theta_Q$ & 0.60 \\
$\theta_L$ & 0.50 & $\theta_M$ & 0.40 \\
$\theta_{K_1}$ & 0.70 & $\theta_{K_2}$ & 0.70 \\
$\theta_{K_1}^{\mathrm{low}}$ & 0.55 & & \\
\bottomrule
\end{tabular}
\end{table}

\subsection{Dependency Constraints}
\label{app:dependency_constraints}

Dependency constraints encode consistency between downstream variables and their
upstream parents. In the medium-complexity regime, the main dependency caps are
specified by the following inequalities:
\[
\begin{aligned}
C_{1,t}
&\leq
0.15+0.45Y_t+0.25P_{m,t}+0.15Q_t+\eta,\\
C_{2,t}
&\leq
0.10+0.35Y_t+0.35P_{b,t}+0.20Q_t+\eta,\\
D_{1,t}
&\leq
0.10+0.40S_{1,t}+0.30C_{1,t}+0.20Y_t+\eta,\\
D_{2,t}
&\leq
0.05+0.25D_{1,t}+0.30S_{2,t}+0.25C_{2,t}+0.20Q_t+\eta,\\
Q_t
&\leq
0.10+0.30Y_t+0.25S_{1,t}+0.25S_{2,t}+0.20D_{1,t}+\eta,\\
K_{1,t}
&\leq
\min(P_{m,t},C_{1,t},D_{1,t},Q_t,L_t,1-M_t)+\eta,\\
K_{2,t}
&\leq
\min(K_{1,t},P_{b,t},C_{2,t},D_{2,t},Q_t,L_t,1-M_t)+\eta .
\end{aligned}
\]
The dependency slack is fixed at $\eta=0.05$. For each cap, the corresponding
violation margin is the positive excess of the generated downstream value above
the right-hand side. These margins are aggregated over all dependency constraints
and time steps to obtain the dependency-family violation score.

\subsection{Mutual-Exclusion Constraints}
\label{app:mutual_exclusion_constraints}

Mutual-exclusion constraints prevent high service outputs from co-occurring with
maintenance or safe-mode conditions. In the medium-complexity regime, the
service variables are constrained as follows:
\[
\begin{aligned}
M_t \geq 0.60 &\quad\Rightarrow\quad K_{1,t} \leq 0.30,\\
M_t \geq 0.60 &\quad\Rightarrow\quad K_{2,t} \leq 0.30.
\end{aligned}
\]
The associated violation score is computed from the positive excess above the
allowed service level when maintenance or safe-mode conditions are active.

\subsection{Temporal-Ordering Constraints}
\label{app:temporal_ordering_constraints}

Temporal-ordering constraints require selected downstream high-service states to be
preceded by stable upstream conditions. The stability window is set to
$q=2$. A high-service state is recorded whenever
$K_{j,t}\geq\theta_{K_j}$, for $j\in\{1,2\}$. For each such state, the corresponding
prerequisites must have remained above their thresholds over the interval
$\{t-q,\ldots,t\}$.

For $K_1$, the prerequisite set is
\[
\mathcal{P}_{K_1}
=
\{P_m,C_1,D_1,Q,L\}.
\]
For $K_2$, the prerequisite set is
\[
\mathcal{P}_{K_2}
=
\{K_1,P_b,C_2,D_2,Q,L\}.
\]
Thus, a temporal-ordering violation is recorded when a high-service state
is observed before the relevant prerequisite variables have remained stable throughout
the required window.

A second temporal condition concerns recovery after cyber degradation. If cyber
integrity falls below $0.35$, then $C_1$, $C_2$, $D_2$, and $Q$ are not allowed
to recover above their high thresholds before $Y$ recovers above $0.50$. The
corresponding violation score aggregates the positive recovery excess observed
before cyber recovery has occurred.

\subsection{Structural Invariants}
\label{app:structural_invariants}

Structural invariants enforce basic graph-state admissibility independently of
scenario-specific service constraints. They include bounded node and edge values,
consistency with the allowed-edge topology, and continuity restrictions over
edge-state evolution. Generated node and allowed-edge values are required to
remain in $[0,1]$. Variables corresponding to forbidden edges are not generated
as independent state components, consistently with the graph-state formulation
in \cref{subsec:methods_dynamic_graphs}.

Continuity restrictions limit abrupt edge-state changes. Unless topology
perturbation is active, the maximum allowed edge jump between consecutive time
steps is
\[
\Delta_{\max}=0.25.
\]
In the medium-complexity regime, the structural-invariant family additionally constrains abrupt recovery of the global service variable $K_2$. For $t\geq 1$, define
\[
\psi_t
=
\mathbf{1}
\left\{
\min
\left(
K_{1,t-1},
D_{2,t-1},
C_{2,t-1},
Q_{t-1}
\right)
<0.55
\right\}
\left[
K_{2,t}-K_{2,t-1}-\Delta_{\max}
\right]_+,
\]
where $[u]_+=\max(u,0)$. Thus, an increase in $K_2$ exceeding $\Delta_{\max}$ contributes to the structural-invariant violation score when at least one of its preceding support variables remains below 0.55.
Structural-invariant violation scores aggregate out-of-range values, allowed-edge inconsistencies, excess edge-state jumps, and, in the medium-complexity regime, the abrupt-recovery term defined above.

\section{Additional Scenario-Validation Analysis} 
\label{app:supplementary_results}

This appendix presents additional analyses supporting the scenario-continuation and structural-validation results reported in Section~\ref{sec:results}.

The additional analyses do not establish separate claims. They provide complementary diagnostics concerning scenario admissibility, ensemble usability, probability calibration, and robustness across the two controlled simulation regimes.

The analyses cover three complementary properties:

\begin{enumerate}
\item structural admissibility of generated scenario continuations;
\item ensemble usability, including retention, effective sample size, and diversity;
\item scenario-level probability calibration.
\end{enumerate}

\subsection{Full Compact-Regime Scenario-Management Metrics}
\label{app:full_strategy_metrics}

The compact graph regime is summarized in \cref{subsec:results_cross_regime},
while the cross-regime summary is reported in
\cref{tab:cross_experiment_results}. \Cref{tab:app_experiment_1_full_metrics}
provides the corresponding full scenario-management metrics for Experiment~1.

\begin{table}[ht]
\caption{Full scenario-management metrics for the compact simulation regime.}
\label{tab:app_experiment_1_full_metrics}
\centering
\footnotesize
\setlength{\tabcolsep}{2.5pt}
\renewcommand{\arraystretch}{1.08}
\begin{tabularx}{\textwidth}{@{}>{\raggedright\arraybackslash}p{0.23\textwidth}
*{7}{>{\centering\arraybackslash}p{0.09\textwidth}}@{}}
\toprule
\textbf{Strategy} &
\shortstack{\textbf{Invalid}\\\textbf{mass}} &
\shortstack{\textbf{Trajectory}\\\textbf{CVR}} &
\shortstack{\textbf{Retained}\\\textbf{ratio}} &
\shortstack{\textbf{ESS}\\\textbf{ratio}} &
\textbf{Diversity} &
\shortstack{\textbf{RS-}\\\textbf{MSE}} &
\shortstack{\textbf{RS-}\\\textbf{ECE}} \\
\midrule
Unconstrained diffusion & 0.002996 & 0.002996 & 1.000000 & 1.000000 & 1.030889 & 0.578380 & 0.741424 \\
Hard filtering & 0.000000 & 0.000000 & 0.997004 & 0.997004 & 1.028912 & 0.577716 & 0.740841 \\
Soft weighting & 0.002715 & 0.002996 & 1.000000 & 0.999930 & 1.030889 & 0.578309 & 0.741360 \\
Optional projection & 0.000000 & 0.000000 & 1.000000 & 1.000000 & 1.030910 & 0.578201 & 0.741282 \\
\bottomrule
\end{tabularx}
\end{table}

\subsection{Soft-Weighting Reduction Summary}
\label{app:soft_weighting_summary}

Soft weighting is discussed in the main text as a probability-reweighting
mechanism rather than as a guarantee of structural admissibility; see
\cref{subsec:results_constraint_handling}. \Cref{tab:app_soft_weighting_effect}
summarizes the reduction in invalid scenario mass and the effective scenario-ensemble size preserved across the two regimes.

\begin{table}[htbp]
\caption{Soft-weighting effect across the two dynamic-graph simulation regimes.}
\label{tab:app_soft_weighting_effect}
\centering
\footnotesize
\setlength{\tabcolsep}{4pt}
\renewcommand{\arraystretch}{1.08}
\begin{tabularx}{\textwidth}{@{}>{\raggedright\arraybackslash}p{0.27\textwidth}
*{4}{>{\centering\arraybackslash}p{0.16\textwidth}}@{}}
\toprule
\textbf{Experiment} &
\shortstack{\textbf{Unconstrained}\\\textbf{invalid mass}} &
\shortstack{\textbf{Soft}\\\textbf{invalid mass}} &
\shortstack{\textbf{Absolute}\\\textbf{reduction}} &
\shortstack{\textbf{ESS}\\\textbf{ratio}} \\
\midrule
Experiment 1: compact & 0.002996 & 0.002715 & 0.000282 & 0.999930 \\
Experiment 2: complex & 0.155929 & 0.148807 & 0.007122 & 0.998764 \\
\bottomrule
\end{tabularx}
\end{table}

\FloatBarrier

\subsection{Perturbation Diagnostics}
\label{app:calibration_robustness_diagnostics}

Robustness diagnostics are discussed in \cref{subsec:results_robustness}.
\Cref{tab:app_robustness_summary} reports the condensed medium-complexity
diagnostics under observation-noise and topology-perturbation conditions.

\begin{table}[ht]
\caption{Condensed scenario-validation robustness diagnostics in the medium-complexity simulation regime.}
\label{tab:app_robustness_summary}
\centering
\footnotesize
\setlength{\tabcolsep}{4pt}
\renewcommand{\arraystretch}{1.08}
\begin{tabularx}{\textwidth}{@{}>{\raggedright\arraybackslash}p{0.27\textwidth}
*{3}{>{\centering\arraybackslash}p{0.22\textwidth}}@{}}
\toprule
\textbf{Condition} &
\shortstack{\textbf{Unconstrained}\\\textbf{invalid mass}} &
\shortstack{\textbf{Hard retained}\\\textbf{ratio}} &
\shortstack{\textbf{Soft invalid}\\\textbf{mass}} \\
\midrule
Noise base & 0.154819 & 0.845181 & 0.147811 \\
Noise severe & 0.158594 & 0.841406 & 0.151528 \\
Topology base & 0.156555 & 0.843445 & 0.149428 \\
Topology severe & 0.169536 & 0.830464 & 0.161635 \\
\bottomrule
\end{tabularx}
\end{table}

\FloatBarrier

\bibliographystyle{unsrtnat}
\bibliography{references}

\end{document}